\documentclass[conference]{IEEEtran}
\usepackage[colorlinks=true, citecolor=blue]{hyperref}
\usepackage{cite}
\usepackage[utf8]{inputenc}
\usepackage{booktabs}   
\usepackage{tabularx}   
\usepackage{caption}   
   \usepackage[numbers]{natbib}
\usepackage{float}  
\usepackage{amsmath,amssymb,amsfonts}
\usepackage{algorithmic}
\usepackage{graphicx}
\usepackage{textcomp}
\usepackage{xcolor}
\usepackage{tabularx}
\usepackage{multirow}  
\usepackage{tikz}
\usetikzlibrary{positioning,arrows.meta,shapes.geometric,fit,backgrounds,calc,shadows}
\definecolor{cData}{HTML}{D7E8FA}   \definecolor{cDataB}{HTML}{2E6DA6}
\definecolor{cPre}{HTML}{DCF0E0}    \definecolor{cPreB}{HTML}{2F8F4E}
\definecolor{cBase}{HTML}{FBE6CF}   \definecolor{cBaseB}{HTML}{C77A21}
\definecolor{cMeta}{HTML}{ECDBF5}   \definecolor{cMetaB}{HTML}{7A3FA0}
\definecolor{cOut}{HTML}{FAD9DD}    \definecolor{cOutB}{HTML}{B23347}
\definecolor{cGroup}{HTML}{F4F6F8}  \definecolor{cArrow}{HTML}{3A3A3A}
\begin{document}

\title{A Deep Learning-Based Stacking Ensemble Framework 
for Turbofan Engine Remaining Useful Life Prediction}

\author{
\IEEEauthorblockN{
Limon Bin Hossain$^{1}$,
Md. Salehin Seyam$^{1}$,
Md Rashedul Islam$^{2}$,
Abdur Rahman$^{3}$,
and Md Sharifuzzaman $^{4}$
}
\IEEEauthorblockA{
$^{1}$Department of Industrial and Production Engineering\\
Bangladesh University of Engineering and Technology\\
Dhaka 1000, Bangladesh
}
\IEEEauthorblockA{
$^{2}$Department of Industrial Engineering\\
Clemson University\\
Clemson, SC, USA
}
\IEEEauthorblockA{
$^{3}$College of Engineering and Science\\
Louisiana Tech University\\
Ruston, LA 71270, USA
}
\IEEEauthorblockA{
$^{4}$School of Engineering Technology\\
Purdue Polytechnic Institute, Purdue University\\
West Lafayette, IN, USA
}
}
\maketitle
\begin{abstract}
This study proposes a two-level stacking ensemble framework for 
Remaining Useful Life (RUL) prediction of turbofan engines, evaluated 
on the NASA C-MAPSS benchmark using the FD001 and FD003 subsets. The 
framework integrates four heterogeneous deep learning base learners Long Short-Term Memory (LSTM), Convolutional Neural Network (CNN), 
CNN--LSTM, and CNN--GRU whose out-of-fold predictions are combined 
by an XGBoost meta-learner to capture complex degradation patterns 
while mitigating individual model biases. Comprehensive experiments 
demonstrate that the stacking ensemble achieves superior predictive 
performance, with Root Mean Square Error (RMSE) of 9.989 and 8.613, 
Mean Absolute Error (MAE) of 7.081 and 5.195, and $R^2$ of 0.899 and 
0.906 for FD001 and FD003, respectively. Compared to the best-reported 
baseline (TCAT: RMSE 11.12 and 11.02), the proposed method achieves 
RMSE reductions of 10.2\% and 21.8\% for FD001 and FD003, 
respectively. Feature correlation analysis, residual diagnostics, and 
training convergence curves validate the model's robustness. These 
findings underscore the efficacy of stacking ensemble methods for 
prognostics and health management in safety-critical aerospace 
applications.
\end{abstract}


\section{Introduction}

Remaining Useful Life (RUL) is the estimated time until a system or 
component fails or exceeds acceptable performance limits. In 
safety-critical applications such as aircraft turbofan and gas turbine 
engines, the accuracy of RUL estimation is closely tied to flight 
safety, operational reliability, and economic efficiency. Inaccurate 
predictions may contribute to unplanned maintenance interventions or, 
in severe cases, accelerated component degradation \cite{lodygowski_unsupervised_2025, 
maulana_explainable_2023}. Benchmark datasets such as NASA's C-MAPSS 
are widely adopted for evaluating prognostic models in aero-engine 
health management \cite{xu_research_2025}. Recent advances reflect 
growing adoption of deep learning, transformer-based architectures, 
hybrid models, digital twins, and uncertainty-aware frameworks to 
improve RUL prediction accuracy in complex degradation systems 
\cite{wahid_self-attention_2023, star_deep_2024}.

RUL prediction extends beyond engines and is essential for aerospace 
composite structures and a broad range of industrial assets. Key 
challenges arise from variability in material properties, operating 
conditions, and fault mechanisms, which complicate degradation 
indicator extraction and uncertainty quantification 
\cite{baptista_self-organizing_2021, dersin_analysis_2026}. 
To address these challenges, this study proposes a stacking ensemble 
of heterogeneous deep learning base learners, fused by an XGBoost 
meta-learner, to deliver high-accuracy RUL prediction competitive 
with the state of the art. The specific research objectives pursued 
in this study are:

\begin{itemize}
  \item \textbf{RO1:} To develop a preprocessing pipeline using 
  sensor selection, min--max normalization, piecewise-linear RUL 
  labeling, and fixed-length sliding windows for C-MAPSS turbofan 
  degradation data.

  \item \textbf{RO2:} To train four heterogeneous deep learning base 
  learners LSTM, CNN, CNN--LSTM, and CNN--GRU for capturing 
  complementary temporal and spatial degradation patterns.

  \item \textbf{RO3:} To construct a two-level stacking ensemble in 
  which an XGBoost meta-learner is trained on out-of-fold predictions 
  of the base learners to improve accuracy and generalization.

  \item \textbf{RO4:} To evaluate the proposed framework on the FD001 
  and FD003 subsets using RMSE, MAE, and $R^2$, supported by 
  correlation, residual, and convergence analyses.
\end{itemize}
\section{Related Work}

Numerous studies have applied machine learning, deep learning, and 
statistical methods to the C-MAPSS benchmark, underscoring the need 
for increasingly accurate prognostic models. Asif et al.\ 
\cite{asif_deep_2022} developed a deep LSTM-based RUL prediction 
framework incorporating data augmentation, normalization, and a 
piecewise-linear degradation model, with hyperparameters optimized 
via iterative grid search, achieving strong performance across all 
four C-MAPSS subsets. The SCTA-LSTM framework \cite{tian_spatial_2023} 
integrates spatial correlation and temporal attention mechanisms, 
reporting RMSE values of 12.10 and 12.14 for FD001 and FD003, 
respectively. Deng et al.\ \cite{deng_prediction_2024} proposed a 
CNN--LSTM--Attention model in which CNN extracts local sensor features, 
LSTM captures temporal dependencies, and an attention mechanism 
redistributes critical information, achieving RMSE values of 15.977 
and 13.907 on FD001 and FD003. 

Xuan et al.\ \cite{xuan_uncertainty-aware_2023} demonstrated that 
ensemble approaches outperform traditional methods, improving both 
RUL accuracy and uncertainty quantification through hyperparameter 
optimization and early stopping. The Two-stream Convolutional 
Augmented Transformer (TCAT) \cite{10272711} currently reports among 
the lowest published RMSE values (11.12 and 11.02) on FD001 and FD003. 
Sharma et al.\ \cite{sharma2024framework} reported that a Light 
Gradient Boosting Machine achieved an RMSE of 7.95 for FD003, while 
Random Forest yielded an RMSE of 11.59 for FD001. Deniz et al.\ 
\cite{deniz2025enhancing} proposed an RMSE-calibrated ensemble 
integrating CatBoost, XGBoost, and Random Forest, achieving an RMSE 
of 13.72 and $R^2$ of 0.8909 on FD001.

Despite these advances, most existing methods rely on single 
architectures or unstructured model combinations, without 
systematically exploiting the complementary strengths of 
heterogeneous deep learners through stacked generalization. 
Furthermore, few studies report MAE and $R^2$ alongside RMSE, 
limiting the completeness of performance evaluation. This study 
addresses these gaps by proposing a structured two-level stacking 
ensemble that combines LSTM, CNN, CNN--LSTM, and CNN--GRU base 
learners via an XGBoost meta-learner, evaluated with a comprehensive 
set of metrics.

\section{Proposed Methodology}

This section formalizes the proposed two-level stacking ensemble. 
We first describe the data representation and preprocessing, 
followed by the heterogeneous deep base learners, the stacked 
generalization procedure with its XGBoost meta-learner, and 
finally the evaluation metrics.
\subsection{Problem Formulation}
Let the multivariate sensor recording of an engine unit be a sequence
$\mathbf{X}=\{\mathbf{x}_1,\mathbf{x}_2,\dots,\mathbf{x}_{T}\}$, where
$\mathbf{x}_t\in\mathbb{R}^{F}$ is the vector of $F$ monitored signals
at operating cycle~$t$ and $T$ is the unit's last observed cycle. The goal of RUL prediction is to learn a mapping that estimates the residual life $\hat{y}$ from a recent window
$\mathbf{W}$ of multivariate measurements, such that $\hat{y}$
approximates the true RUL $y$ at the last cycle of the window.
\begin{equation}
\mathcal{F}:\ \mathbb{R}^{L_w\times F}\ \longrightarrow\ \mathbb{R}_{\ge 0},
\qquad \hat{y}=\mathcal{F}(\mathbf{W}),
\label{eq:mapping}
\end{equation}
\subsection{Data Preprocessing}

\subsubsection{Sensor Selection}
The C-MAPSS recordings contain $21$ sensor channels and $3$ operational
settings. For the FD001 and FD003 subsets, which operate under a single
condition, several sensors remain constant or exhibit negligible
variance throughout the engine's life and therefore carry no degradation
information. These flat channels (sensors $1,5,6,10,16,18,19$) are
discarded, retaining the $F=14$ informative sensors
$\{2,3,4,7,8,9,11,12,13,14,15,17,20,21\}$. This selection is supported
by the correlation analysis in Section ~\ref{sec:results}, where the Retained sensors show a strong monotonic association with the
degradation trend.

\subsubsection{Min--Max Normalization}
Because the retained sensors span heterogeneous physical units and
magnitudes, each feature $j$ is linearly rescaled to the $[0,1]$
interval.
\begin{equation}
\tilde{x}^{(j)}_t=\frac{x^{(j)}_t-x^{(j)}_{\min}}
{x^{(j)}_{\max}-x^{(j)}_{\min}},
\qquad j=1,\dots,F,
\label{eq:minmax}
\end{equation}
where $x^{(j)}_{\min}$ and $x^{(j)}_{\max}$ are the minimum and maximum
of sensor~$j$ estimated on the training set only and then
applied unchanged to the test set, preventing information leakage.

\subsubsection{Piecewise-Linear RUL Target}
Assuming a constant RUL during the early, healthy phase of operation
and a linear decay thereafter is more physically faithful than a purely
linear label. We therefore cap the target at a maximum useful life
$R_{\max}=125$ cycles.
\begin{equation}
y_t=\min\!\left(R_{\max},\, T-t\right).
\label{eq:pwlin}
\end{equation}
This piecewise-linear formulation discourages the model from
extrapolating implausibly large RUL values when the engine is far from
failure and concentrates learning capacity on the informative
degradation region.

\subsubsection{Sliding-Window Representation}
Deep sequence models require fixed-length inputs that preserve temporal
context. A window of length $L_w=40$ is slid over each normalized
engine sequence with unit stride, producing samples each paired with the label $y_{i+L_w-1}$ at the window's last cycle, as
defined in~\eqref{eq:pwlin}. A length of $40$ cycles offers a favorable
trade-off, providing sufficient temporal memory to characterize the
degradation slope while retaining enough training windows from
shorter-lived units. During inference, a single window comprising the final $L_w$ 
cycles of each test unit is constructed and used for prediction.
\begin{equation}
\mathbf{W}_i=\big[\tilde{\mathbf{x}}_{i},\tilde{\mathbf{x}}_{i+1},
\dots,\tilde{\mathbf{x}}_{i+L_w-1}\big]\in\mathbb{R}^{L_w\times F},
\label{eq:window}
\end{equation}
\subsection{Base Learners}
Four complementary architectures make up the base-learner pool. Each
is trained independently to minimize the mean squared error on the
windowed samples, so that their prediction errors are partially
decorrelated, and thus amenable to ensemble fusion.

\subsubsection{Long Short-Term Memory (LSTM) Architecture}
The LSTM models long-range temporal dependencies via gated memory
cells. At step~$t$, given input $\mathbf{x}_t$ and previous hidden
state $\mathbf{h}_{t-1}$, the gates are
\begin{align}
\mathbf{f}_t&=\sigma\!\left(\mathbf{W}_f[\mathbf{h}_{t-1},\mathbf{x}_t]+\mathbf{b}_f\right),\\
\mathbf{i}_t&=\sigma\!\left(\mathbf{W}_i[\mathbf{h}_{t-1},\mathbf{x}_t]+\mathbf{b}_i\right),\\
\tilde{\mathbf{C}}_t&=\tanh\!\left(\mathbf{W}_C[\mathbf{h}_{t-1},\mathbf{x}_t]+\mathbf{b}_C\right),\\
\mathbf{C}_t&=\mathbf{f}_t\odot\mathbf{C}_{t-1}+\mathbf{i}_t\odot\tilde{\mathbf{C}}_t,\\
\mathbf{o}_t&=\sigma\!\left(\mathbf{W}_o[\mathbf{h}_{t-1},\mathbf{x}_t]+\mathbf{b}_o\right),\\
\mathbf{h}_t&=\mathbf{o}_t\odot\tanh\!\left(\mathbf{C}_t\right),
\end{align}
where $\sigma(\cdot)$ is the logistic sigmoid, $\odot$ denotes
element-wise multiplication, and $\mathbf{f}_t,\mathbf{i}_t,\mathbf{o}_t$
are the forget, input, and output gates. Two stacked LSTM layers are
followed by a dropout and a fully connected regression head that maps the
final hidden state to a scalar RUL.

\subsubsection{Convolutional Neural Network (CNN) Architecture}
The one-dimensional CNN extracts local degradation patterns by sliding
learnable kernels along the time axis. For the $k$-th feature map of
layer~$l$,
\begin{equation}
\mathbf{z}^{(l)}_{k}=\phi\!\left(\sum_{c}\mathbf{w}^{(l)}_{k,c}\ast
\mathbf{z}^{(l-1)}_{c}+b^{(l)}_{k}\right),
\label{eq:conv}
\end{equation}
where $\ast$ is the 1-D convolution, $\phi(\cdot)$ is the ReLU
activation, and $c$ indexes input channels. Stacked convolution--pooling
blocks are flattened and passed to a dense layer for RUL regression.

\subsubsection{CNN--LSTM Architecture}
This hybrid first applies convolutional blocks to the window to distil
salient local features, and then feeds the resulting feature sequence
to an LSTM that models the temporal evolution of those features:
\begin{equation}
\hat{y}=\mathrm{FC}\big(\mathrm{LSTM}(\mathrm{CNN}(\mathbf{W}))\big).
\label{eq:cnnlstm}
\end{equation}
The CNN reduces noise and dimensionality while the LSTM captures
long-term degradation trends, combining spatial and temporal modeling.

\subsubsection{CNN--GRU Architecture}
The second hybrid replaces the recurrent block with a Gated Recurrent
Unit (GRU), a lighter gated architecture with comparable temporal
modeling capacity. Its update and reset gates are
\begin{align}
\mathbf{z}_t&=\sigma\!\left(\mathbf{W}_z[\mathbf{h}_{t-1},\mathbf{x}_t]+\mathbf{b}_z\right),\\
\mathbf{r}_t&=\sigma\!\left(\mathbf{W}_r[\mathbf{h}_{t-1},\mathbf{x}_t]+\mathbf{b}_r\right),\\
\tilde{\mathbf{h}}_t&=\tanh\!\left(\mathbf{W}_h[\mathbf{r}_t\odot\mathbf{h}_{t-1},\mathbf{x}_t]+\mathbf{b}_h\right),\\
\mathbf{h}_t&=(1-\mathbf{z}_t)\odot\mathbf{h}_{t-1}+\mathbf{z}_t\odot\tilde{\mathbf{h}}_t,
\end{align}
where $\mathbf{z}_t$ and $\mathbf{r}_t$ are the update and reset gates.
The GRU's fewer parameters reduce the overfitting risk on the limited
training windows while retaining sequence-modeling power.

\subsection{Stacking Ensemble}
\label{sec:stacking}
The proposed framework follows stacked generalization, in which a
\emph{meta-learner} is trained on the predictions of the base learners
rather than on the raw inputs. To obtain unbiased meta-features and
avoid target leakage, the base-model predictions are generated through
$K$-fold cross-validation ($K=5$).

The windowed training set $\mathcal{D}=\{(\mathbf{W}_i,y_i)\}_{i=1}^{N}$
is partitioned into $K$ disjoint folds
$\mathcal{D}_1,\dots,\mathcal{D}_K$. For each base model
$m\in\{1,\dots,M\}$ (here $M=4$) and each fold $k$, the model is trained
on $\mathcal{D}\setminus\mathcal{D}_k$ and used to predict the held-out
fold. The resulting \emph{out-of-fold} (OOF) prediction for sample~$i$
in fold~$k$ is
\begin{equation}
z^{(m)}_i=f^{(-k)}_m(\mathbf{W}_i),\qquad i\in\mathcal{D}_k,
\label{eq:oof}
\end{equation}
where $f^{(-k)}_m$ denotes base model~$m$ fitted without fold~$k$.
Stacking the OOF predictions of all base learners yields the
meta-feature vector
\begin{equation}
\mathbf{z}_i=\big[z^{(1)}_i,z^{(2)}_i,\dots,z^{(M)}_i\big]\in\mathbb{R}^{M},
\label{eq:metafeat}
\end{equation}
and the meta-dataset
$\mathcal{D}_{\text{meta}}=\{(\mathbf{z}_i,y_i)\}_{i=1}^{N}$. Each base
model is additionally retrained on the \emph{full} training set to
produce predictions for the unseen test windows, which are assembled
into test-time meta-features in the same manner.

An XGBoost regressor $g(\cdot)$ is trained on
$\mathcal{D}_{\text{meta}}$ to produce the final estimate
$\hat{y}_i=g(\mathbf{z}_i)$. XGBoost builds an additive ensemble of
$B$ regression trees,
\begin{equation}
\hat{y}_i=\sum_{b=1}^{B}\eta\,f_b(\mathbf{z}_i),\qquad f_b\in\mathcal{T},
\label{eq:xgbadd}
\end{equation}
where $\mathcal{T}$ is the space of CART trees and $\eta$ is the
learning rate. The trees are grown by minimizing the regularized
objective
\begin{equation}
\mathcal{L}=\sum_{i=1}^{N}\ell\!\left(y_i,\hat{y}_i\right)
+\sum_{b=1}^{B}\Omega(f_b),\quad
\Omega(f)=\gamma\,T_{\!f}+\tfrac{1}{2}\lambda
\lVert\mathbf{w}\rVert^{2},
\label{eq:xgbobj}
\end{equation}
in which $\ell$ is the squared-error loss, $\mathcal{T}_{\!f}$ is the
number of leaves, $\mathbf{w}$ are the leaf weights, and
$\gamma,\lambda$ control tree complexity and $L_2$ regularization. At
boosting round~$t$ the objective is approximated to second order using
the gradients $g_i=\partial_{\hat{y}}\ell$ and Hessians
$h_i=\partial^2_{\hat{y}}\ell$, giving the optimal weight of leaf~$j$
\begin{equation}
w_j^{\ast}=-\frac{\sum_{i\in I_j}g_i}{\sum_{i\in I_j}h_i+\lambda},
\label{eq:leafweight}
\end{equation}
where $I_j$ is the set of samples routed to leaf~$j$. By learning a non-linear, regularized combination of the base 
predictions, the meta-learner exploits their complementary 
strengths and suppresses the idiosyncratic errors of any 
single model.

\subsection{Evaluation Metrics}
Model performance is assessed with three standard regression metrics.
The Root Mean Square Error penalizes large deviations,
\begin{equation}
\text{RMSE}=\sqrt{\frac{1}{n}\sum_{i=1}^{n}\left(y_i-\hat{y}_i\right)^2},
\label{eq:rmse}
\end{equation}
The Mean Absolute Error measures average deviation magnitude,
\begin{equation}
\text{MAE}=\frac{1}{n}\sum_{i=1}^{n}\left|y_i-\hat{y}_i\right|,
\label{eq:mae}
\end{equation}
and the coefficient of determination quantifies explained variance,
\begin{equation}
R^{2}=1-\frac{\sum_{i=1}^{n}\left(y_i-\hat{y}_i\right)^2}
{\sum_{i=1}^{n}\left(y_i-\bar{y}\right)^2},
\label{eq:r2}
\end{equation}
where $\bar{y}$ is the mean of the true RUL. Lower RMSE and MAE, combined with a higher $R^{2}$, 
indicate better predictive performance.

\subsection{Training Configuration}
All deep base learners are trained with the Adam optimizer 
(initial learning rate $10^{-3}$) by minimizing the mean 
squared error, using a batch size of $256$ for up to $50$ 
epochs with early stopping (patience = 10) on a held-out 
validation split (10\% of training data) to prevent 
overfitting; a dropout rate of $0.2$ is applied after 
recurrent and dense layers for regularization. The XGBoost
meta-learner uses squared-error loss with a learning rate $\eta=0.05$,
maximum tree depth of $4$, subsample and column-subsample ratios of
$0.8$, and $L_2$ regularization $\lambda=1$, with the number of trees
selected by early stopping. These settings, together with the
five-fold OOF protocol, ensure that the reported test metrics reflect
genuine generalization rather than memorization. 

\section{Dataset and Experimental Setup}

All experiments utilize the NASA Commercial Modular 
Aero-Propulsion System Simulation (C-MAPSS) benchmark 
\cite{4711414}. C-MAPSS models the degradation of a 
high-bypass turbofan engine by integrating thermodynamic 
engine models with actuator and sensor noise to produce 
high-fidelity multivariate time-series data. The dataset 
comprises four subsets (FD001--FD004) that vary in the 
number of operating conditions and fault modes. This study 
focuses on FD001 and FD003, both operating under a single 
condition, to isolate the effect of fault mode complexity. 
Table~\ref{tab:cmapss_stats} summarizes the structural 
characteristics of all four subsets. All experiments were 
implemented in Python using TensorFlow/Keras for the deep 
base learners and the XGBoost library for the meta-learner.
\begin{table}[!ht]
\centering
\caption{Structural statistics of the NASA C-MAPSS benchmark subsets}

\renewcommand{\arraystretch}{1.3}
\begin{tabular}{lcccc}
\toprule
\textbf{Property} & \textbf{FD001} & \textbf{FD002} & \textbf{FD003} & \textbf{FD004} \\
\midrule
Training Engines      & 100    & 260    & 100    & 249    \\
Test Engines          & 100    & 259    & 100    & 248    \\
Training Samples      & 17,731 & 48,558 & 21,120 & 56,815 \\
Test Samples          & 100    & 259    & 100    & 248    \\
Max Cycle (Train)     & 362    & 378    & 525    & 543    \\
Min Cycle (Train)     & 128    & 128    & 145    & 128    \\
Max Cycle (Test)      & 303    & 367    & 475    & 486    \\
Min Cycle (Test)      & 31     & 21     & 38     & 19     \\
Operating Conditions  & 1      & 6      & 1      & 6      \\
Fault Modes           & 1      & 1      & 2      & 2      \\
\bottomrule
\end{tabular}
\label{tab:cmapss_stats}
\end{table}

\section{Results and Discussion}
\label{sec:results}

The predictive performance of the proposed stacking ensemble was rigorously evaluated against established baselines and contemporary deep learning architectures, including LSTM, CNN, CNN-LSTM, and CNN-GRU, using the NASA C-MAPSS benchmark. Table~\ref{tab:wide_comparison} presents a comparative analysis against state-of-the-art (SOTA) approaches for the FD001 and FD003 subsets. The stacking ensemble achieved RMSE values of 9.989 and 8.613 for FD001 and FD003, respectively, representing a substantial improvement over the TCAT model (11.12, 11.02)~\cite{10272711} and the SCTA-LSTM model (12.10, 12.14)~\cite{tian_spatial_2023}. The proposed method also outperforms the CNN-LSTM-Attention model (15.977, 13.907)~\cite{deng_prediction_2024} and the k-LSTM-GFT framework (13.10, 11.27)~\cite{nunes_combining_2025}. The Transformer-based approach~\cite{elkawakjy_transformer-based_2025} exhibited considerably higher error rates (33.60, 35.49), suggesting that attention mechanisms alone may be insufficient to capture degradation patterns without appropriate architectural constraints.

\begin{table*}[tbp]
\centering
\caption{Comparison of RUL Prediction Performance on FD001 and FD003 Datasets}
\label{tab:wide_comparison}
\small
\begin{tabular*}{\textwidth}{@{\extracolsep{\fill}}lcccccc}
\toprule
\textbf{Methodology} & \multicolumn{3}{c}{\textbf{FD001}} & \multicolumn{3}{c}{\textbf{FD003}} \\
\cmidrule{2-4} \cmidrule{5-7}
& \textbf{RMSE} & \textbf{MAE} & \textbf{R$^2$} & \textbf{RMSE} & \textbf{MAE} & \textbf{R$^2$} \\
\midrule
Simple Ageing Model~\cite{sanchez_simplified_2023}        & 16.79  & -- & -- & 17.48  & -- & -- \\
PILmin-maxregret~\cite{sanchez_physics-informed_2023}      & 15.11  & -- & -- & 16.99  & -- & -- \\
CNN-LSTM-Attention~\cite{deng_prediction_2024}            & 15.977 & -- & -- & 13.907 & -- & -- \\
k-LSTM-GFT~\cite{nunes_combining_2025}                   & 13.10  & -- & -- & 11.27  & -- & -- \\
TCAT~\cite{10272711}                                      & 11.12  & -- & -- & 11.02  & -- & -- \\
SCTA-LSTM~\cite{tian_spatial_2023}                        & 12.10  & -- & -- & 12.14  & -- & -- \\
Transformer-based~\cite{elkawakjy_transformer-based_2025} & 33.60  & -- & -- & 35.49  & -- & -- \\
\midrule
\textbf{Proposed Method (Stacking Ensemble)} & \textbf{9.989} & \textbf{7.081} & \textbf{0.899} & \textbf{8.613} & \textbf{5.195} & \textbf{0.906} \\
\bottomrule
\end{tabular*}
\smallskip
\raggedright
\footnotesize
TCAT = Two-stream Convolutional Augmented Transformer~\cite{10272711};
k-LSTM-GFT = k-Long Short-Term Memory with Gated Feature Transformation~\cite{nunes_combining_2025}.
\end{table*}

Table~\ref{tab:single_metrics} provides a performance comparison of the proposed ensemble and the baseline models. For FD001, the stacking ensemble achieved RMSE of 9.989, MAE of 7.081, and $R^2$ of 0.899, substantially outperforming LSTM (12.93, 9.657, 0.831), CNN (14.986, 11.268, 0.773), CNN-LSTM (13.968, 11.162, 0.802), and CNN-GRU (13.071, 10.121, 0.827). Similar improvements were observed for FD003, with the ensemble achieving RMSE of 8.613, MAE of 5.195, and $R^2$ of 0.906, compared to LSTM (11.189, 7.704, 0.841). The superior $R^2$ values indicate that the ensemble explains approximately 90\% of variance in RUL predictions.

\begin{table}[tbp]
\centering
\caption{Performance Metrics for Baseline and Proposed Models}
\label{tab:single_metrics}
\footnotesize
\begin{tabularx}{\columnwidth}{X c c c}
\toprule
\textbf{Model} & \textbf{RMSE} & \textbf{MAE} & \textbf{R$^2$} \\
\midrule
\textit{Dataset: FD001} & & & \\
LSTM     & 12.93  & 9.657  & 0.831 \\
CNN      & 14.986 & 11.268 & 0.773 \\
CNN-LSTM & 13.968 & 11.162 & 0.802 \\
CNN-GRU  & 13.071 & 10.121 & 0.827 \\
\midrule
\textbf{Stacking Ensemble} & \textbf{9.989} & \textbf{7.081} & \textbf{0.899} \\
\midrule
\textit{Dataset: FD003} & & & \\
LSTM     & 11.189 & 7.704  & 0.841 \\
CNN      & 14.655 & 12.418 & 0.727 \\
CNN-LSTM & 13.746 & 10.278 & 0.760 \\
CNN-GRU  & 15.448 & 12.107 & 0.697 \\
\midrule
\textbf{Stacking Ensemble} & \textbf{8.613} & \textbf{5.195} & \textbf{0.906} \\
\bottomrule
\end{tabularx}
\end{table}

\begin{figure}[tbp]
    \centering
    \includegraphics[width=\columnwidth]{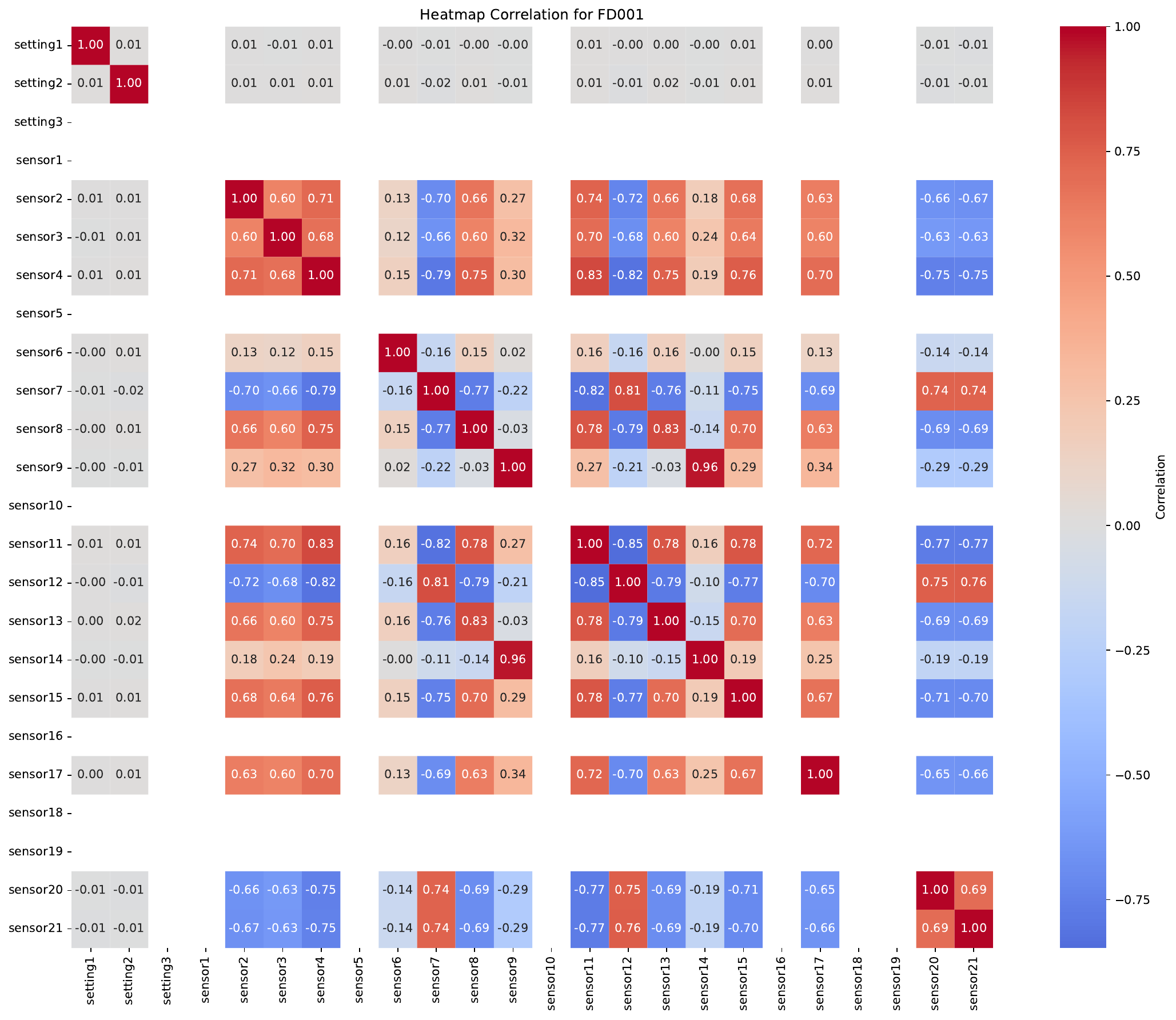}
    \caption{Correlation heatmap for the FD001 subset.}
    \label{fig:heatmap1}
\end{figure}

\begin{figure}[tbp]
    \centering
    \includegraphics[width=\columnwidth]{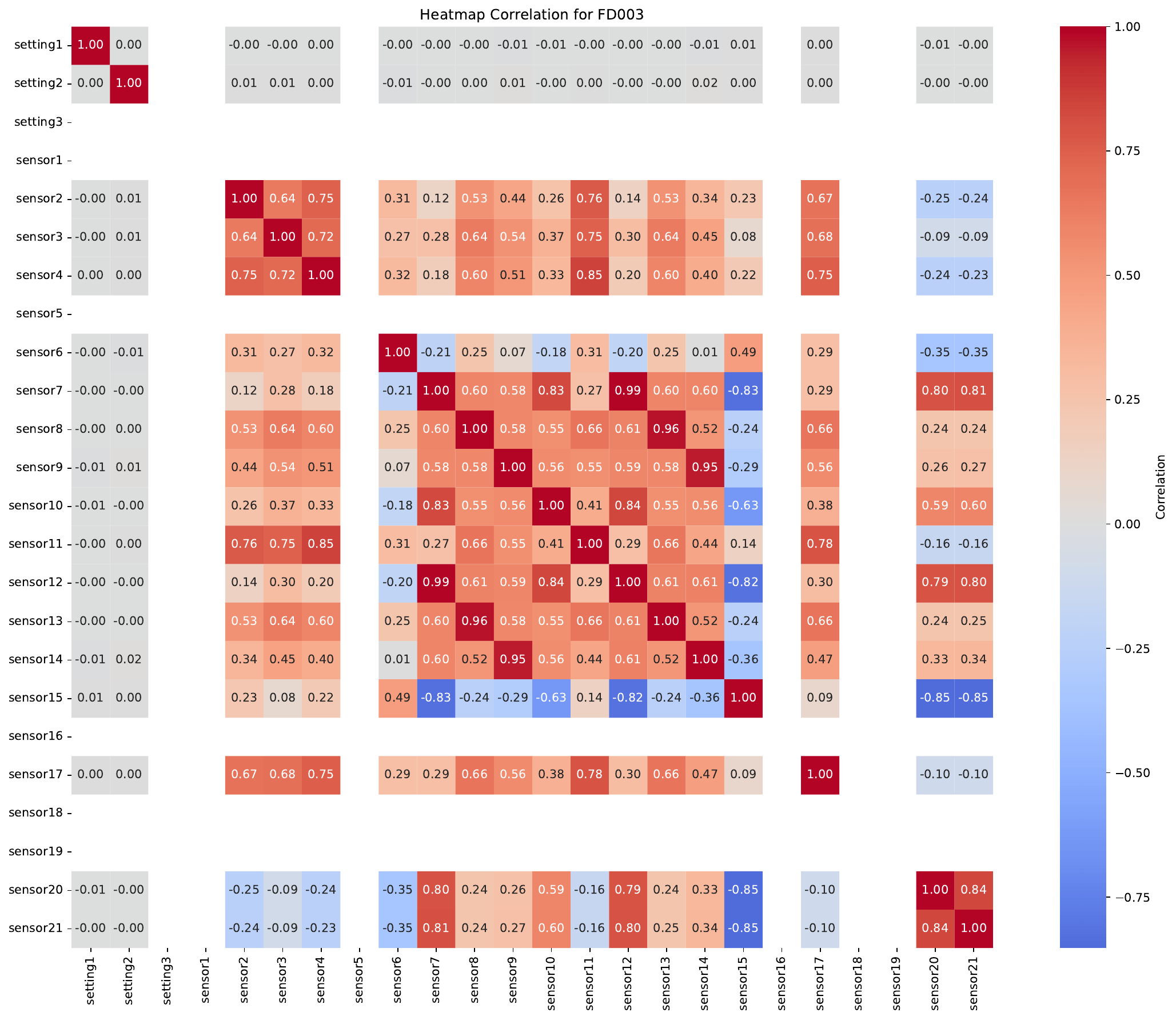}
    \caption{Correlation heatmap for the FD003 subset.}
    \label{fig:heatmap2}
\end{figure}

\begin{figure}[tbp]
    \centering
    \includegraphics[width=\columnwidth]{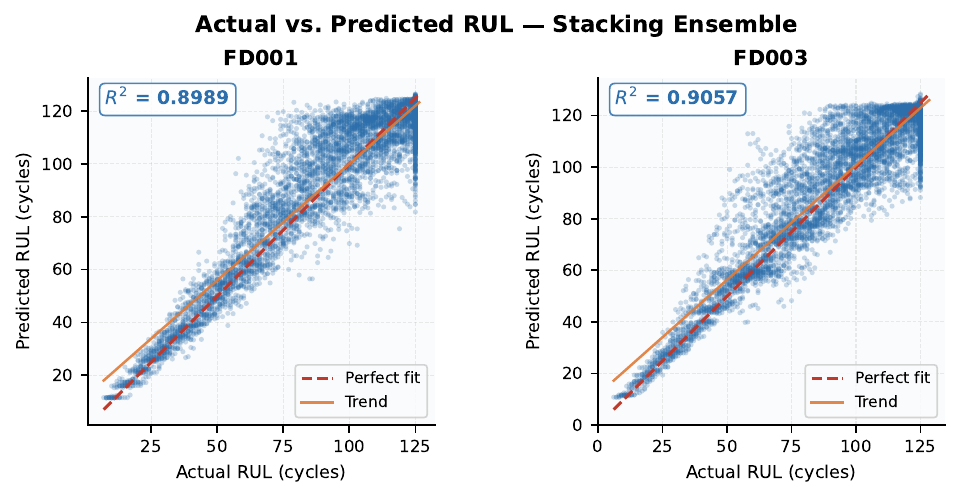}
    \caption{Actual vs.\ predicted RUL for FD001 and FD003.}
    \label{fig:actual_vs_predicted}
\end{figure}

Feature correlation analysis, illustrated in Figures~\ref{fig:heatmap1} and~\ref{fig:heatmap2}, reveals distinct correlation structures between FD001 and FD003. Sensors 12, 7, and 15 exhibit strong correlations with operational settings, informing feature selection for model development. The actual versus predicted RUL plots (Figure~\ref{fig:actual_vs_predicted}) demonstrate excellent agreement with the ideal fit line, with $R^2$ values of 0.899 and 0.906 for FD001 and FD003, respectively, corroborating the ensemble's predictive reliability.

\begin{figure}[tbp]
    \centering
    \includegraphics[width=\columnwidth]{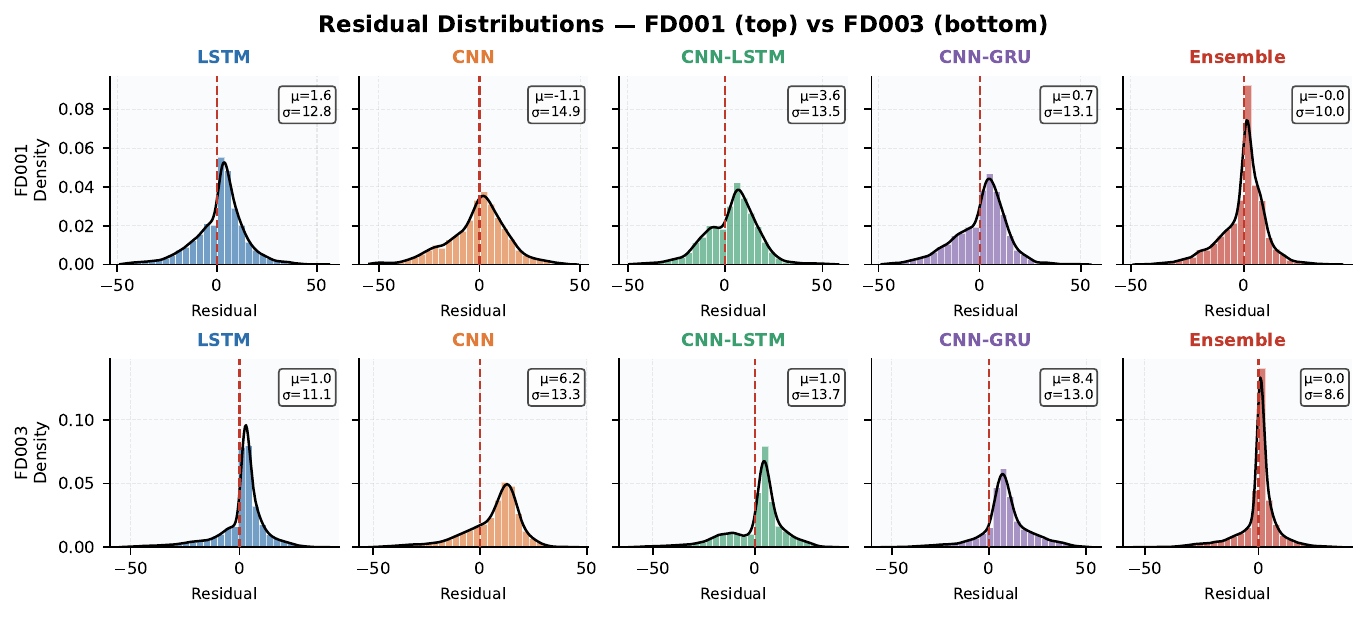}
    \caption{Residual analysis of the stacking ensemble predictions.}
    \label{fig:Residual_Analysis}
\end{figure}

\begin{figure}[tbp]
    \centering
    \includegraphics[width=\columnwidth]{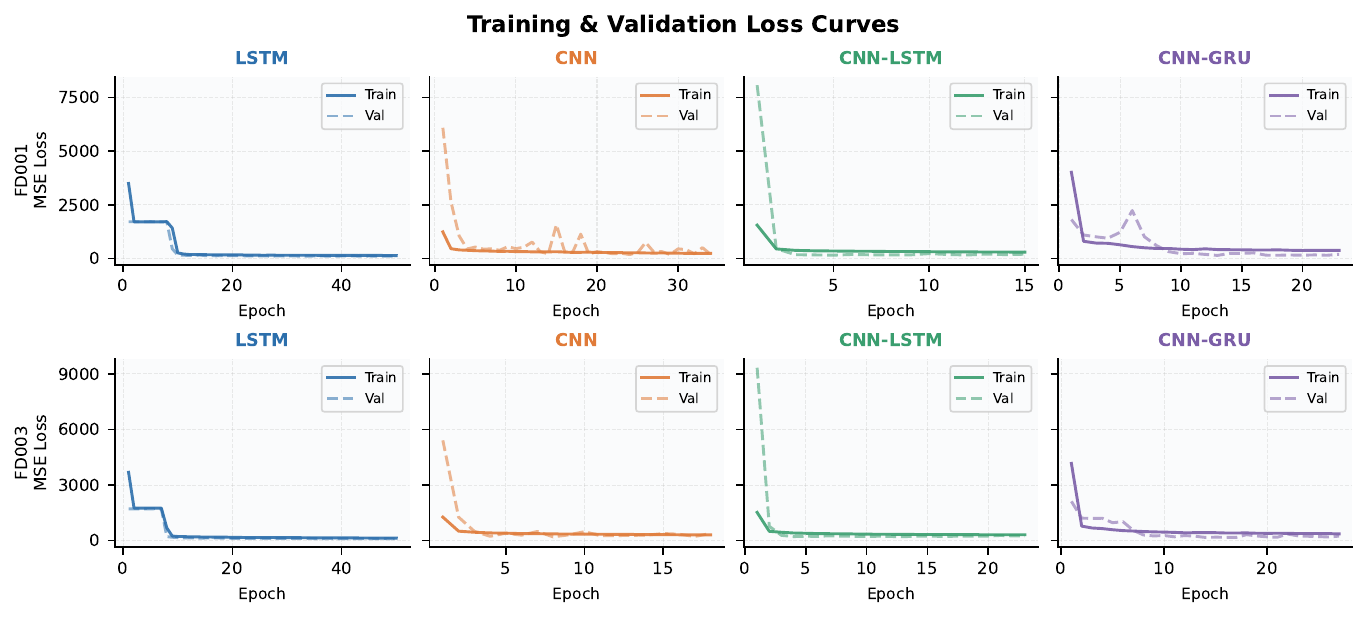}
    \caption{Training convergence curves of the stacking ensemble.}
    \label{fig:Training_Curves}
\end{figure}

\begin{figure}[tbp]
    \centering
    \includegraphics[width=\columnwidth]{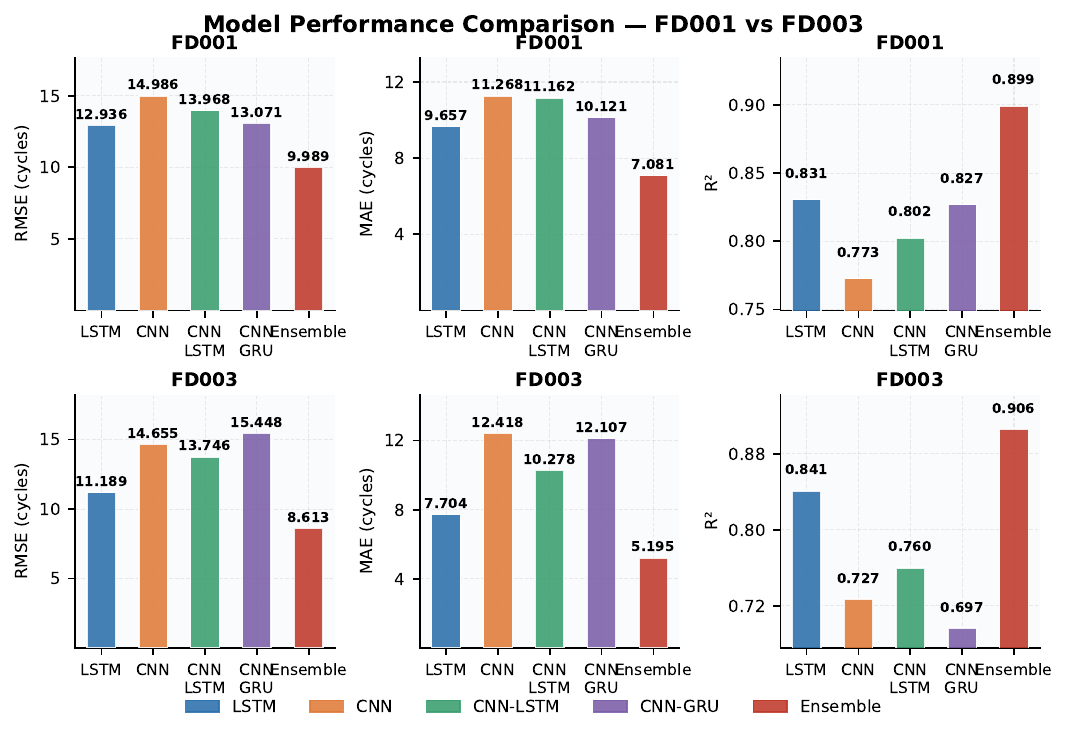}
    \caption{Performance comparison across all evaluated models.}
    \label{fig:Model_Comparison}
\end{figure}

Residual analysis (Figure~\ref{fig:Residual_Analysis}) confirms homoscedasticity and normally distributed errors with minimal bias, validating the model assumptions. Training curves (Figure~\ref{fig:Training_Curves}) exhibit smooth convergence without overfitting, indicating effective regularization within the ensemble architecture. Figure~\ref{fig:Model_Comparison} provides a visual comparison of model performance, clearly demonstrating the stacking ensemble's superiority across all evaluated metrics.

\section{Conclusion and Future Work}

\subsection{Conclusion}

This study proposed a two-level stacking ensemble framework 
for turbofan engine RUL prediction and evaluated it against 
LSTM, CNN, CNN--LSTM, and CNN--GRU baselines on the NASA 
C-MAPSS benchmark. Focusing on the FD001 and FD003 subsets, 
the proposed model achieved RMSE of 9.989, MAE of 7.081, 
and $R^2$ of 89.9\% for FD001, and RMSE of 8.613, MAE of 
5.195, and $R^2$ of 90.6\% for FD003. These results 
represent improvements of 10.2\% and 21.8\% in RMSE over 
the previously best-reported TCAT baseline (11.12 and 11.02) 
for FD001 and FD003, respectively. The ensemble also reports 
the lowest MAE and highest $R^2$ values among the compared 
methods, validating the effectiveness of stacked 
generalization for prognostics and health management in 
aerospace applications.

\subsection{Future Work}

Future research should explore the integration of attention 
mechanisms and transformer blocks as additional base learners 
within the stacking framework, potentially further improving 
predictive accuracy. Extending the evaluation to the 
multi-condition FD002 and FD004 subsets, as well as 
real-world industrial datasets, would strengthen the 
generalizability of the findings. Additionally, domain 
adaptation strategies and uncertainty quantification 
techniques warrant investigation to enhance model 
deployability in operational settings with limited 
labeled data.

\bibliographystyle{IEEEtran}
\bibliography{ref}
\end{document}